\documentclass[runningheads]{llncs}
\usepackage{graphicx}
\usepackage{amsmath,amssymb} 
\usepackage{color}
\usepackage{booktabs}
\usepackage[width=122mm,left=12mm,paperwidth=146mm,height=193mm,top=12mm,paperheight=217mm]{geometry}
\usepackage[pagebackref=false,breaklinks=true,letterpaper=true,colorlinks,bookmarks=false,linkcolor=blue,anchorcolor=blue,citecolor=blue]{hyperref}
\usepackage{expl3}
\usepackage{algorithm}
\usepackage{algpseudocode}

\ExplSyntaxOn
\newcommand\latinabbrev[1]{
  \peek_meaning:NTF . {
    #1\@}%
  { \peek_catcode:NTF a {
      #1.\@ }%
    {#1.\@}}}
\ExplSyntaxOff

\def\ie{\latinabbrev{\emph{i.e}}}

\begin{document}
\pagestyle{headings}
\mainmatter
\title{POI: Multiple Object Tracking with High Performance Detection and Appearance Feature} 

\titlerunning{POI}

\authorrunning{Fengwei Yu, Wenbo Li, Quanquan Li, Yu Liu, Xiaohua Shi, Junejie Yan}

\author{$\textrm{Fengwei Yu}^{1,3} \quad \textrm{Wenbo Li}^{2,3} \quad \textrm{Quanquan Li}^{3} \quad \textrm{Yu Liu}^{3}$\\ $\textrm{Xiaohua Shi}^{1} \quad \textrm{Junjie Yan}^{3}$}

\institute{$\textrm{}^{1}$Beihang University, China \\ $\textrm{}^{2}$University at Albany, SUNY, USA\\ $\textrm{}^{3}$Sensetime Group Limited, China}

\maketitle

\begin{abstract}
Detection and learning based appearance feature play the central role in data association based multiple object tracking (MOT), but most recent MOT works usually ignore them and only focus on the hand-crafted feature and association algorithms. In this paper, we explore the high-performance detection and deep learning based appearance feature, and show that they lead to significantly better MOT results in both online and offline setting. We make our detection and appearance feature publicly available\footnote{\url{https://drive.google.com/open?id=0B5ACiy41McAHMjczS2p0dFg3emM}}. In the following part, we first summarize the detection and appearance feature, and then introduce our tracker named Person of Interest (POI), which has both online and offline version\footnote{We use POI to denote our online tracker and KDNT to denote our offline tracker in submission.}.
\end{abstract}

\section{Detection}
In data association based MOT, the tracking performance is heavily affected by the detection results. We implement our detector based on Faster R-CNN \cite{RenHGS15}. In our implementation, the CNN model is fine-tuned from the VGG-16 on ImageNet. The additional training data includes ETHZ pedestrian dataset \cite{EssLSG08}, Caltech pedestrian dataset \cite{DollarWSP09} and the self-collected surveillance dataset (365653 boxes in 47556 frames). We adopt the multi-scale training strategy by randomly sampling a pyramid scale for each time. However, we only use a single scale and a single model during test. Moreover, we also use skip pooling \cite{BellZBG15} and multi-region \cite{GidarisK15} strategies to combine features at different scales and levels.

In considering the definition of MOTA in MOT16 \cite{MilanL0RS16}, the sum of false negatives (FN) and false positives (FP) poses a large impact on the value of MOTA. In Table~\ref{tab0}, we show that our detection optimization strategies lead to the significant decrease in the sum of FP and FN\footnote{We use detection score threshold 0.3 for Faster R-CNN and -1 for DPMv5 , labeling the ID of detection box with incremental integer, and evaluate FP and FN with MOT16 devkit.}.
\begin{table}\scriptsize
\renewcommand\arraystretch{1.5}
\caption{Detection Performance Evaluation(on MOT16 train set)}
\centering
\begin{tabular}{|p{14em}*{4}{c}|}
 \hline
 \makebox[14em]{Strategies } & \makebox[3em]{FP} & \makebox[3em]{FN} & \makebox[3em]{FP+FN}  &\\
  \hline
 \makebox[14em]{DPMv5 } & \makebox[3em]{28839} & \makebox[3em]{62353} & \makebox[3em]{91192}  &\\
 \makebox[14em]{Faster R-CNN baseline } & \makebox[3em]{5384} & \makebox[3em]{47343} & \makebox[3em]{52727}  &\\
 \makebox[14em]{Faster R-CNN + skip pooling } & \makebox[3em]{5410} & \makebox[3em]{46399} & \makebox[3em]{51809}  &\\
 \makebox[14em]{Faster R-CNN + multi-region } & \makebox[3em]{4476} & \makebox[3em]{46738} & \makebox[3em]{51214}  &\\
  \makebox[14em]{Faster R-CNN + both } & \makebox[3em]{8722} & \makebox[3em]{37865} & \makebox[3em]{46587}  &\\
 \hline
\end{tabular}
\label{tab0}
\end{table}

\section{Appearance Feature}
The distance between appearance features is used for computing the affinity value in data association. The affinity value based on the ideal appearance feature should be large for persons of the same identity, and be small for persons of different identities. In our implementation, we extract the appearance feature using a network which is similar to GoogLeNet \cite{SzegedyLJSRAEVR15}. The input size of our network is $96 \times 96$, and the kernel size of $pool5$ layer is $3 \times 3$ instead of $7 \times 7$. The output layer is a fully connected layer which outputs the 128 dimensional feature. In the tracking phase, patches are first cropped according to the detection responses, and then resized to $96 \times 96$ for feature extraction. The cosine distance is used for measuring the appearance affinity.

For training, we collect a dataset which contains nearly 119 K patches from 19835 identities. Such a dataset consists of multiple person re-id datasets, including PRW \cite{ZhengZSCT16}, Market-1501 \cite{ZhengZSCT16}, VIPeR \cite{ProsserZGX10} and CUHK03 \cite{LiZXW14}. We use the softmax and triplet loss jointly during training. The softmax loss guarantees the discriminative ability of the appearance feature, while the triplet loss ensures the cosine distance of the appearance features of the same identity to be small.

\section{Online Tracker}
We implement a simple online tracker, which uses Kalman filter \cite{kalman1960new} for motion prediction and Kuhn-Munkres algorithm \cite{kuhn1955hungarian} for data association. The overall tracking procedure is described in Algorithm~\ref{alg:Alg1}.

\begin{algorithm}
\begin{algorithmic}[1]
\scriptsize
\Require A new frame at the $t$-th timestep, the detection set $D^{t}$, and the tracklet set $T^{t-1}$
\Ensure The new tracklet set $T^{t}$
\State Calculate the affinity matrix $A^{t-1} = \textit{Affinity}(T^{t-1},D^{t})$
\State Divide $T^{t-1}$ into high tracking quality set $T^{t-1}_{high}$ and low quality set $T^{t-1}_{low}$ with threshold $\tau_{t}$
\State Use Kuhn-Munkres algorithm to find the optimal matching between $(T^{t-1}_{high}, T^{t-1}_{low})$ and $D^{t}$ based on $A^{t-1}$
\State Use threshold $\tau_{a}$ to decide whether association success or not
\State Obtain association-success set $T^{t-1}_{success_{i}}$ with matched detection set $D^{t}_{success_{i}}$, association-fail tracklet set $T^{t-1}_{fail}$ and unmatched detection set $D^{t}_{fail}$
\State Use Kalman filter and feature aggregation to generate new tracklet subset $T^{t}_{1}$ based on association-success set: $T^{t}_{1} = Average(T^{t-1}_{success_{i}},D^{t}_{success_{i}})$.
\State Use Kalman filter to predict or remove the association-fail tracklets with missing tracklets threshold $\tau_{m}$: $T^{t}_{2} = Predict\_Or\_Remove(T^{t-1}_{fail},\tau_{m})$.
\State Initialize the unmatched detections as the new tracklets: $T^{t}_{3} = Initialize(D^{t}_{fail})$.
\State Merge the tracklet subsets to generate new candidate tracklet  set : $T^{t}_{candidate} = T^{t}_{1} \cup T^{t}_{2} \cup T^{t}_{3} $.
\State Remove out of image border candidate tracklet set to generate new tracklet set: $T^{t} = Filter(T^{t}_{candidate})$
\end{algorithmic}
\caption{Overall Procedure of the Online Tracker}
\label{alg:Alg1}
\end{algorithm}

In the following, we introduce the affinity matrix construction, data association method, threshold value setting and tracking quality metric.

\noindent\textbf{Affinity Matrix Construction.}
To construct an affinity matrix for the Kuhn-Munkres algorithm, we calculate the affinity between tracklets and detections. We combine motion, shape and appearance affinity as the final affinity. Specifically, the appearance affinity is calculated based on the appearance feature described in Section 2. Details of the affinity calculation are given below:
\vspace{-2mm}
\begin{equation}
    \textit{aff}_{app}(trk_{i},det_{j}) = cosine(feat_{trk_{i}},feat_{det_{j}})
\end{equation}
\vspace{-2mm}
\begin{equation}
    \textit{aff}_{mot}(trk_{i},det_{j}) = e^{-w_1 * ((\frac{X_{trk_{i}}-X_{det_{j}}}{W_{det_{j}}})^2+(\frac{Y_{trk_{i}}-Y_{det_{j}}}{H_{det_{j}}})^2)}
\end{equation}
\vspace{-2mm}
\begin{equation}
    \textit{aff}_{shp}(trk_{i},det_{j}) = e^{-w_2 * (\frac{|H_{trk_{i}}-H_{det_{j}}|}{H_{trk_{i}}+H_{det_{j}}} + \frac{|W_{trk_{i}}-W_{det_{j}}|}{W_{trk_{i}}+W_{det_{j}}})}
\end{equation}
\vspace{-2mm}
\begin{equation}
    \textit{affinity}(trk_{i},det_{j}) = \textit{aff}_{app}(trk_{i},det_{j}) *  \textit{aff}_{mot}(trk_{i},det_{j}) * \textit{aff}_{shp}(trk_{i},det_{j})
\end{equation}
$\textit{aff}_{app}$, $\textit{aff}_{mot}$ and $\textit{aff}_{shp}$ indicate appearance, motion and shape affinity between the detection and tracklet, respectively. We combine these affinities with weights $w_1$ and $w_2$ as the final affinity.

\noindent\textbf{Data Association.}
The tracklets and new detections are associated using the Kuhn-Munkres algorithm. Since the Kuhn-Munkres algorithm attempts to yield the global optimal result, it may fail when some detections are missing. To this end, we use a two-stage matching strategy, which divides $T^{t-1}$ into high tracking quality set $T^{t-1}_{high}$ and low quality set $T^{t-1}_{low}$. The matching is first performed between $T^{t-1}_{high}$ and $D$, and then performed between $(T^{t-1}_{high} - T^{t-1}_{success}) \cup T^{t-1}_{low}$ and $D-D_{success}$.

\noindent\textbf{Threshold Value Setting.}
On line 2 of Algorithm~\ref{alg:Alg1}, we introduce $\tau_{t}$  to divide $T^{t-1}$ into high and low tracking quality set. The strategy is intuitive: we mark a tracklet with \emph{high} flag whose tracking quality is higher than $\tau_{t}$, other tracklets will be mark as \emph{low}. On line 4, we use $\tau_{a}$ to mark the association to be success or fail based on the affinity value. On line 7, we use $\tau_{m}$ as a threshold to drop a tracklet which is lost for more than $\tau_{m}$ frames.

\noindent\textbf{Tracking Quality Metric.}
Tracking quality is designed to measure whether a object is tracking well or not. We use following formula to define tracking quality:
\begin{equation}
 Quality(tracklet_i) = \frac{\sum_{k \in couples(tracklet_i)}^{}{\textit{affinity}_k}}{length(tracklet_i)}(1 - e^{-w_3 * \sqrt{length(tracklet_i)}})
\end{equation}
where $couples(tracklet_i)$, with the form $\{trk_x,det_y\}$, is a set that contains every success association couple in history.
\section{Offline Tracker}
Our offine tracker an improved version of H$^2$T \cite{WenLYLYL14} while based on K-Dense Neighbors \cite{LiuYLY12}. It is more robust and efficient than H$^2$T in handling the complex tracking scenarios. The overall procedure of the tracker is described in Algorithm~\ref{alg:Alg2}.

\vspace{-2mm}
\begin{algorithm}
\begin{algorithmic}[1]
\scriptsize
\Require A tracking video and the detections in all frames
\Ensure The tracking results (trajectories of targets)
\State Divide the tracking video into multiple disjoint segments in the temporal domain
\State Use the dense neighbors (DN) search\footnotemark to associate the detection responses into short tracklets in each segment
\While{The number of segments is greater than one}{}
\State Merge several nearby segments into a longer segment
\State Use the DN search in each longer segment to associate existing tracklets into longer tracklets
\EndWhile
\end{algorithmic}
\caption{Overall Procedure of the Offline Tracker}
\label{alg:Alg2}
\end{algorithm}
\footnotetext{The DN search is performed on an affinity matrix which encodes the similarity between two tracklets. Please refer to \cite{DuQLWHL16,LiWCZLL15,LiuYLY12,WenLYLYL14} for details about DN search and its advantages over the GMCP \cite{LiWCL15,ZamirDS12} as a data association method.}
\vspace{-2mm}

We make the following improvements over H$^2$T \cite{WenLYLYL14}.

\noindent\textbf{Appearance Representation.}
To construct the affinity matrix for the dense neighbors (DN) search, we need to calculate three affinities, \ie, appearance, motion, and smoothness affinity. Among these three affinities, the appearance affinity is the most important one and we use the CNN based feature described in Section. 2, instead of the hand-crafted feature in \cite{WenLYLYL14}.

\noindent\textbf{Big Target.}
A scenario that H$^2$T \cite{WenLYLYL14} does not work well is the mixture of small and big targets. The reason is that the motion and smoothness affinities are unreliable for the big targets. Such unreliability is caused by the unsteady detection responses of the big targets. We introduce two thresholds, $\tau_{s}$ and $\tau_{r}$, regarding the object scale to deal with this challenge, \ie, $\tau_{s}$ for preventing associating detection responses from very different scale, and $\tau_{r}$ for determining whether to reduce the weights of motion and smoothness affinity. Specifically, if the ratio of the detection response scale and the target scale is less than $\tau_{s}$, such a detection response will not be associated with the target. If the ratio of the detection response height and the image height is greater than $\tau_{r}$, such a detection response will not be associated with the target. Both $\tau_{s}$ and $\tau_{r}$ are set as 0.5.

\noindent\textbf{Algorithm Efficiency.}
H$^2$T is slow in handling the long tracking sequence where there exist plenty of targets. Among the steps in the algorithm, the step of DN search is the most time-consuming. To be more specific, the larger an affinity matrix, the longer time it will take to perform the DN search. Thus, we abandon the high-order information \cite{WenLYLYL14} when constructing the affinity matrix, which significantly reduces the matrix dimensions and improves the algorithm efficiency.

\vspace{-4mm}

\section{Evaluation}
Our online and offline tracker are not learning based algorithm. We only tuning detection score threshold on train set and apply it to its similar scene from test set. For evaluation and submission, 0.1 is set for MOT16-03 and MOT16-04 due to high precision of detection result (03 and 04 are both surveillance scene, which is quite easy while our detector have been trained by self-collected surveillance dataset), and 0.3 is set for other sequences.

For both online\footnote{we use 0.5 for $w_1$, 1.5 for $w_2$, 1.2 for $w_3$,0.5 for $\tau_{t}$, 0.4 for $\tau_{a}$ and 100 frames for $\tau_{m}$.}  and offline tracker, we compare our detector with the official detector, and compare our feature with default CNN feature. The comparison results on MOT16 \cite{MilanL0RS16} train set are listed in Table~\ref{tab1} and Table~\ref{tab2}, respectively. Note that our detector leads to much better results in MT, ML, FP and FN, and our feature helps reduce both IDS and FM.

\begin{table}\scriptsize
\renewcommand\arraystretch{1.5}
\caption{Online Tracker Result On the Train Set}
\centering
\begin{tabular}{|p{13em}*{8}{c}|}\hline
 \makebox[13em]{Det. and Feat.} & \makebox[3.5em]{MT} & \makebox[3.5em]{ML} & \makebox[3em]{FP} & \makebox[3em]{FN} & \makebox[3em]{IDS} & \makebox[3em]{FM} & \makebox[3em]{MOTA} & \makebox[3em]{MOTP}\\\hline
 \makebox[13em]{DPMv5 + Our Feat.} & \makebox[3.5em]{7.54\%} & \makebox[3.5em]{52.42\%} & \makebox[3em]{6197} & \makebox[3em]{70952} & \makebox[3em]{784} & \makebox[3em]{2697} & \makebox[3em]{29.4} & \makebox[3em]{77.2}\\
 \makebox[13em]{Our Det. + GoogLeNet Feat.} & \makebox[3.5em]{31.72\%} & \makebox[3.5em]{16.25\%} & \makebox[3em]{\textbf{3207}} & \makebox[3em]{35472} & \makebox[3em]{1541} & \makebox[3em]{2235} & \makebox[3em]{63.6} & \makebox[3em]{\textbf{82.6}}\\
 \makebox[13em]{Our Det. and Feat.} & \makebox[3.5em]{\textbf{37.33\%}} & \makebox[3.5em]{\textbf{14.70\%}} & \makebox[3em]{3497} & \makebox[3em]{\textbf{34241}} & \makebox[3em]{\textbf{716}} & \makebox[3em]{\textbf{1973}} & \makebox[3em]{\textbf{65.2}} & \makebox[3em]{82.4}\\\hline
\end{tabular}
\label{tab1}
\end{table}
\vspace{-4mm}
\begin{table}\scriptsize
\renewcommand\arraystretch{1.5}
\caption{Offline Tracker Result On the Train Set}
\centering
\begin{tabular}{|p{13em}*{8}{c}|}\hline
 \makebox[13em]{Det. and Feat.} & \makebox[3.5em]{MT} & \makebox[3.5em]{ML} & \makebox[3em]{FP} & \makebox[3em]{FN} & \makebox[3em]{IDS} & \makebox[3em]{FM} & \makebox[3em]{MOTA} & \makebox[3em]{MOTP}\\\hline
 \makebox[13em]{DPMv5 + Our Feat.} & \makebox[3.5em]{10.64\%} & \makebox[3.5em]{52.80\%} & \makebox[3em]{27238} & \makebox[3em]{63443} & \makebox[3em]{1540} & \makebox[3em]{1853} & \makebox[3em]{16.5} & \makebox[3em]{77.4}\\
 \makebox[13em]{Our Det. + GoogLeNet Feat.} & \makebox[3em]{13.93\%} & \makebox[3.5em]{60.93\%} & \makebox[3em]{\textbf{1258}} & \makebox[3em]{58213} & \makebox[3em]{1350} & \makebox[3em]{2196} & \makebox[3em]{44.9} & \makebox[3em]{\textbf{85.0}}\\
 \makebox[13em]{Our Det. and Feat.} & \makebox[3.5em]{\textbf{37.52\%}} & \makebox[3.5em]{\textbf{17.60\%}} & \makebox[3em]{2762} & \makebox[3em]{\textbf{33327}} & \makebox[3em]{\textbf{462}} & \makebox[3em]{\textbf{717}} & \makebox[3em]{\textbf{66.9}} & \makebox[3em]{83.3}\\\hline
\end{tabular}
\label{tab2}
\end{table}

\section{ECCV 2016 Challenge Results.}
\begin{table}\scriptsize
\renewcommand\arraystretch{1.5}
\caption{Comparison to the State-of-the-art Methods On MOT16 Rank List}
\centering
\begin{tabular}{|p{13em}*{8}{c}|}\hline

 \makebox[13em]{Tracker} & \makebox[3.5em]{MT} & \makebox[3.5em]{ML} & \makebox[3em]{FP} & \makebox[3em]{FN} & \makebox[3em]{IDS} & \makebox[3em]{FM} & \makebox[3em]{MOTA} & \makebox[3em]{MOTP}\\\hline

 \makebox[13em]{KFILDAwSDP (Online)} & \makebox[3.5em]{26.9\%} & \makebox[3.5em]{21.6\%} & \makebox[3em]{23266} & \makebox[3em]{56394} & \makebox[3em]{1977} & \makebox[3em]{2954} & \makebox[3em]{55.2} &\makebox[3em]{77.2}\\
 \makebox[13em]{MCMOT-HDM (Offline)} & \makebox[3.5em]{31.5\%} & \makebox[3.5em]{24.2\%} & \makebox[3em]{9855} & \makebox[3em]{57257} & \makebox[3em]{1394} & \makebox[3em]{1318} & \makebox[3em]{62.4} & \makebox[3em]{78.3}\\\hline

 \makebox[13em]{Our Online Tracker} & \makebox[3.5em]{33.99\%} & \makebox[3.5em]{20.82\%} & \makebox[3em]{\textbf{5061}} & \makebox[3em]{55914} & \makebox[3em]{\textbf{805}} & \makebox[3em]{3093} & \makebox[3em]{66.1} & \makebox[3em]{\textbf{79.5}}\\
 \makebox[13em]{Our Offline Tracker} & \makebox[3.5em]{\textbf{40.97}\%} & \makebox[3.5em]{\textbf{18.97}\%} & \makebox[3em]{11479} & \makebox[3em]{\textbf{45605}} & \makebox[3em]{933} & \makebox[3em]{\textbf{1093}} & \makebox[3em]{\textbf{68.2}} & \makebox[3em]{79.4}\\\hline

\end{tabular}
\label{tab3}
\end{table}

Our ECCV 2016 Challenge results are listed in Table~\ref{tab3}. Obviously, both our online and offline trackers outperform the state-of-the-art approaches by a large margin. Note that our offline tracker achieves the best performance in FN. However, its performance in FP is moderate, due to the interpolation module.

\section{Conclusion}

In this submission, we take many efforts to get high performance detection and deep learning based appearance feature. We show that they lead to the state-of-the-art multiple object tracking results, even with very simple online tracker. One observation is that with high performance detection and appearance feature, the state-of-the-art offline tracker does not have expected advantages over the much simpler online one. This observation is not reported in many current MOT papers, which often use detections that are not good enough. We make our detections and deep learning based re-ID features on MOT2016 publicly available, and hope that they can help more sophisticated trackers to get better performance.

\bibliographystyle{splncs03}
\bibliography{egbib}
\end{document}